\documentclass[journal]{IEEEtran}
\usepackage{color}
\usepackage{array}
\usepackage{prettyref}
\usepackage{booktabs}
\usepackage{url}
\usepackage{multirow}
\usepackage{amsthm}
\usepackage{amssymb}
\usepackage{graphicx}
\theoremstyle{plain}
\newtheorem{thm}{\protect\theoremname}
\theoremstyle{plain}
\newtheorem{prop}[thm]{\protect\propositionname}

\providecommand{\propositionname}{Proposition}
\providecommand{\theoremname}{Assumption}

\usepackage{cite}

\ifCLASSINFOpdf

  \DeclareGraphicsExtensions{.pdf,.jpeg,.png,.eps}
\else
  \DeclareGraphicsExtensions{.eps}
\fi
\usepackage[cmex10]{amsmath}
\usepackage{array}

\ifCLASSOPTIONcompsoc
  \usepackage[caption=false,font=normalsize,labelfont=sf,textfont=sf]{subfig}
\else
  \usepackage[caption=false,font=footnotesize]{subfig}
\fi
\usepackage{amssymb}

\begin{document}

%
\title{Voronoi-based Compact Image Descriptors: Efficient Region-of-Interest Retrieval With \\VLAD\ And Deep-Learning--based Descriptors }
%
%
%

\author{Aaron~Chadha and Yiannis Andreopoulos
\thanks{The authors are with the Electronic and Electrical Engineering Department, University College London, Roberts Building, Torrington Place, London, WC1E 7JE, UK (e-mail:
\{aaron.chadha.14, i.andreopoulos\}@ucl.ac.uk). This work has been presented in part at the 2015 Int. Conf. on Computer Vision Syst. (ICVS 2015). The authors acknowledge the support from Innovate UK (project VideoClarity, 101932) and EPSRC (project EP/M00113X/1 and EPSRC CASE award co-sponsored by BAFTA). A. Chadha was supported in part by a fellowship from the Royal Commission from the Exhibition of 1851.}
}

%
%

\markboth{}%
{Shell \MakeLowercase{\textit{et al.}}: Bare Demo of IEEEtran.cls for Journals}
%



\maketitle

\begin{abstract}
We investigate the problem of image retrieval based on visual queries when the latter comprise arbitrary regions-of-interest (ROI) rather than entire images. Our proposal is a compact image descriptor that combines the state-of-the-art in content-based descriptor extraction with a multi-level, Voronoi-based spatial partitioning of each dataset image. The proposed multi-level Voronoi-based encoding uses a spatial hierarchical K-means over interest-point locations, and computes a content-based descriptor\ over each cell. In order to reduce the matching complexity with minimal or no sacrifice in retrieval performance: \textit{(i)} we utilize the tree structure of the spatial hierarchical K-means to perform a top-to-bottom pruning for local similarity maxima; \textit{(ii)} we propose a new image similarity score that combines relevant information from all partition levels into a single measure for similarity; \textit{(iii)} we combine our proposal with a novel and efficient approach for optimal bit allocation within quantized descriptor representations. By deriving  both a Voronoi-based VLAD descriptor (termed as Fast-VVLAD) and a Voronoi-based deep convolutional neural network (CNN) descriptor (termed as Fast-VDCNN), we demonstrate that our Voronoi-based framework is agnostic to the descriptor basis, and can easily be slotted into existing frameworks. Via a range of ROI queries in two standard datasets, it is shown that the Voronoi-based descriptors achieve  comparable or higher mean Average Precision against conventional grid-based  spatial search, while offering more than two-fold reduction in complexity. Finally, beyond ROI queries, we show that Voronoi partitioning improves the geometric invariance of compact CNN descriptors, thereby resulting in competitive performance to the current state-of-the-art on whole image retrieval. \     \ \ \ \ 
\end{abstract}


%
\IEEEpeerreviewmaketitle

\section{Introduction}
\label{sec:intro}

\IEEEPARstart{I}{mage} retrieval based on visual queries is a topic of intensive research interest since it finds many applications in visual search \cite{arandjelovic2012three,bai2016smooth, chadha2015region}, detection of copyright violations \cite{liu2013near}, recommendation services \cite{xu2014obsir} and object or person identification \cite{zheng2015scalable}. For much of the past decade, the state-of-the-art for content-based image retrieval was  to encode the image by  first describing salient points  using a locally-invariant feature descriptor, such as SIFT \cite{lowe2004distinctive} or an image decomposition (e.g., wavelets \cite{spiliotopoulos2001quantization,andreopoulos2008incremental,andreopoulos2002new,andreopoulos2001local}).  As such, a visual vocabulary is  learned offline using K-means or mixture-of-Gaussians (MoG) clustering \cite{kontorinis2009statistical}, which quantizes the feature space into cells (visual words).   The SIFT cell assignments of each database (or query) image are then produced and aggregated in order to obtain a compact representation that can be used for visual-query based retrieval. Notable contributions in this domain have relied on the bag-of-words (BoW) image representation \cite{sivic2003video}, where the SIFTs assigned to each visual word are aggregated into a histogram used for retrieval purposes. Amongst the successful extensions to BoW are feature soft-assignment \cite{philbin2008lost}, spatial matching methods \cite{philbin2007object, gonzalez2014generative, zhong2015fast} and indexing methods \cite{sivic2003video, zheng2015fast}.  

Despite the success of BoW approaches, their large storage and memory access requirements make them unsuitable for image retrieval within large image datasets (e.g., tens of millions of images). For such problems,  the vector of locally aggregated descriptors (VLAD)  \cite{jegou2010aggregating} was introduced, as a non-probabilistic variant of the Fisher vector image descriptor \cite{perronnin2007fisher} that encodes the distribution of SIFT assignments according to cluster centers. VLAD has been shown to achieve very competitive retrieval performance to BoW methods with orders-of-magnitude reduction in complexity and memory footprint, i.e., requiring 16--256 bytes per image instead of  the tens-of-kilobytes required by BoW methods  \cite{jegou2010aggregating}. With such a reduced memory footprint, it has been shown that a standard multicore server can load and retain the VLADs of\ a billion-image dataset in its random access memory \cite{arandjelovic2013all}. This facilitates the scale-up of visual search to big data by using standard cloud computing clusters comprising groups of tens or even hundreds of such servers \cite{lu2015distributed}. 
  
  However, with increasing dataset sizes and complexity in
retrieval, there is a need for deeper models to learn more complex representations and with larger learning capacity. To this end, deep convolutional neural networks (CNNs) have recently come to the forefront  in visual recognition \cite{girshick2014rich,sermanet2013overfeat,karpathy2014large,krizhevsky2012imagenet}. Deep CNNs, as well as hybrid neural network variants like  the FV-NN approach of Peronnin \textit{et al.} \cite{perronnin2015fisher}, have the potential to go beyond  ``shallow" learned encodings like VLAD because they are scalable with training. For example, deep CNNs trained discriminatively on a large and diverse labelled dataset like ImageNet \cite{deng2009imagenet} have been shown to outperform Fisher vectors for image classification \cite{krizhevsky2012imagenet}. In addition, recent work \cite{babenko2014neural, bai2016smooth, azizpour2014factors, yosinski2014transferable} demonstrates that features extracted from intermediate layers of a deep CNN  are actually transferable to image retrieval. The aggregated features tend to provide a rich semantic representation of the image, which in the case of retrieval, has been shown to offer comparable, if not substantially better performance to VLAD and Fisher vector descriptors \cite{babenko2014neural, razavian2014cnn, gong2014multi}. However, deep CNNs are not without some disadvantages: beyond the high computational cost that is inherent with a large training set and numerous layers \cite{krizhevsky2012imagenet, simonyan2014very}, the CNN\    activations lack geometric invariance \cite{gong2014multi, chandrasekhar2015practical}, which pertains to the VLAD\ descriptor and its variants remaining a viable option, especially for fine-grained search.\  \

In this paper, we are interested in the problem of designing a visual-query based retrieval system that is capable of handling both small and large-size ``object'',  or, more broadly, region-of-interest (ROI)  queries over  image datasets. Given a ROI representing a visual query, the proposed system should return all images from the database containing this query, with matching complexity and storage requirements that remain of the order of standard encodings.  This is considerably more challenging than whole-image retrieval systems, as the query object may be occluded or distorted, or be seen from different viewpoints and distances in relevant images \cite{philbin2007object}. This is also the reason why the original VLAD proposal does not perform as well for this problem  \cite{arandjelovic2013all}. We therefore propose  a  new Voronoi-based encoding (VE),  in which we spatially partition the image, using a hierarchical K-means, into Voronoi cells and thus compute multiple descriptors over cells.
    We couple this with an adaptive search algorithm that minimizes the overall computation for similarity identification by  first finding the cells most representative to the query and then deriving a novel single-score metric for the  image over these cells. We propose a novel product quantization framework (based on symmetric distance computation) for our proposal. Finally, we show that our proposed framework is agnostic to the descriptor basis by testing  on both a Voronoi-based VLAD descriptor and Voronoi-based deep CNN feature   descriptor  and assessing performance against their respective state-of-the-art variants. Overall, our system design   for object retrieval adheres to the following principles:

\begin{enumerate}
\item
The system should provide for substantial improvement over the base descriptor's (VLAD or CNN)\ mean Average Precision (mAP) when ROI queries are small relative to the image size.

\item
The system should maintain competitive mAP to the base descriptor representations under ROI queries occupying a sizeable proportion (or the entirety) of images.

\item

The system should be amenable to big-data processing, i.e., its descriptors' size and matching complexity should be comparable to the base descriptor. 

\end{enumerate}

In the following section we discuss the background and related work, with Table \ref{tab:Notation-table.1} summarizing the nomenclature.  In Section \ref{sec:VLAD} we present the offline and online components of our proposed system and Section \ref{sec:PQ} presents the extension of the proposed approach to quantized representations.  Section \ref{sec:Experiments} presents experimental results for Fast-VDCNN on  the Holidays dataset \cite{jegou2008hamming} and Fast-VVLAD on the Caltech Cars (Rear) dataset \cite{fergus2003object}, and Section \ref{sec:Conclusion} draws concluding remarks.

\section{Background and Related Work}
\label{sec:background}

\subsection{Vector of Locally Aggregated Descriptors} \label{VLADdesc}
VLAD is a fixed-size compact image representation that stores first-order information associated with clusters of image salient points  \cite{jegou2010aggregating,jegou2012aggregating}. In essence, VLAD\ is intrinsically related to the Fisher vector image descriptor  \cite{perronnin2010large}. 

In the offline part of the VLAD\ encoding, based on a training set of \(D_\text{SIFT}\)-dimensional SIFT descriptors derived from $Y$ training images, a visual word vocabulary is first learned using K-means clustering. This vocabulary comprises {\(K\)} clusters with \(D_{\text{SIFT}}\)-dimensional centroids \(\mathbf{ M} = \{\boldsymbol{\mu}_{1}, \boldsymbol\mu_{2},\dots,\boldsymbol{\mu}_{K}\}\). 

For each new test image $I$ (out of a test dataset comprising $W$ images), \(N\) interest points are  detected (using an affine invariant detector) and described using \(D_{\text{SIFT}}\)-dimensional SIFT descriptors, thus forming a descriptor ensemble   {\(\mathbf{X} = \{{\boldsymbol x}_{1}, \boldsymbol x_{2},\dots,\boldsymbol x_{N}\}\)}.  The descriptors \( \boldsymbol x_n\), $1 \leq n \leq N$, are assigned to the nearest cluster in the vocabulary via a cluster assignment function  \(f(\boldsymbol x_n)\). VLAD then stores the residuals of the SIFT assignments from their associated centroids.  The VLAD \(D_{\text{SIFT}}\)-dimensional  encoding \(\boldsymbol v_{k}\) for the \(k\)-th cluster, $1 \leq k \leq K$, is given by \cite{jegou2010aggregating,jegou2012aggregating}:

\begin{equation} 
\boldsymbol {v}_{k}=\sum_{\forall x_n:f(x_n )= k}{\left({\boldsymbol x}_{n}-{\boldsymbol\mu}_{k}\right)}.
\end{equation}

The VLAD encodings for each cluster are concatenated into a single descriptor \(\boldsymbol{\phi}(I) = \left[\boldsymbol{v}_1, \dots,\boldsymbol{v}_K \right]^\mathrm{T}\) with fixed dimension \(KD_\text{SIFT}\), which is independent of the number of the SIFT descriptors found in the image.
The VLAD vectors are then sign square-rooted and    $L_2$-normalized \cite{jegou2012aggregating} and the  vectors across all $W$ images of the test dataset are thus aggregated into a single $KD_{\text{SIFT}} \times W$ matrix \( \mathbf{\Phi} = \left[ \boldsymbol{\phi}_1, \dots,\boldsymbol{\phi}_W \right]\).  \ 

In a practical system, the SIFT descriptor length {{\(D_{\text{SIFT}}\)}} is typically 128; if the feature space is coarsely quantized with \(K\) set to 64, then the VLAD image descriptor has 8192 dimensions.  Further dimensionality reduction is achieved with principal component analysis (PCA)  (learned on an independent training set), thus further minimizing the  memory footprint per image descriptor  \cite{jegou2012negative,chum2010unsupervised}. The \(KD_{\text{SIFT}} \times D\) projection  matrix $\mathbf{R}$ used by VLAD\ comprises   only the \(D\) largest eigenvectors of the covariance matrix  \cite{jegou2012negative,chum2010unsupervised}. The projected VLAD, \(\widetilde{\boldsymbol{\phi}}_{\text{test}}\), of each image in the test dataset is then $L_2$-normalized, thereby completing the offline part of the VLAD generation. 

During online ROI-query based retrieval, after the VLAD\ encoding and projection of the ROI\ query has been carried out, the similarity  between that and the  (projected)\ VLAD of a test dataset image, $\widetilde{\boldsymbol{\phi}}_\text{ROI}$ and $\widetilde{\boldsymbol{\phi}}_\text{test}$, can be  simply  measured using the squared Euclidean distance \cite{jegou2012aggregating}. With $L_2$ normalized vectors, this is a  monotonic function of the inner product, such that: 

\begin{equation}
S_\text{ROI,test} =\left\langle \widetilde{\boldsymbol{\phi}}_\text{ROI}, \widetilde{\boldsymbol{\phi}}_\text{test} \right\rangle  
\label{similarity},\end{equation}where the similarity score $S_{\text{ROI,test}}$ ranges between -1 (completely dissimilar) to 1 (perfect match).

\begin{table}
\noindent \centering{}\protect\caption{\label{tab:Notation-table.1}Nomenclature Table.}
\begin{tabular}[t]{>{\centering}p{0.2\columnwidth}>{\raggedright}p{0.7\columnwidth}}
\hline 
\noalign{\vskip\doublerulesep}
\multicolumn{1}{c}{\textbf{Symbol }} & \multicolumn{1}{c}{\textbf{Definition}}
\tabularnewline[\doublerulesep]
\hline 
\noalign{\vskip\doublerulesep}
\centering{}$ U$ & dimensions of unprojected descriptor
\tabularnewline[\doublerulesep]
$D,D'$ & dimensions of PCA-projected and truncated descriptor and descriptor blocks (resp.)
\tabularnewline
\noalign{\vskip\doublerulesep}
\centering{}$Y$ & number of training-set images
\tabularnewline
\noalign{\vskip\doublerulesep}
\centering{}$W$ & number of test-dataset images
\tabularnewline[\doublerulesep]
\noalign{\vskip\doublerulesep}
\centering{}$\mathbf{R}$ &  $U \times D$ PCA projection matrix\tabularnewline
\noalign{\vskip\doublerulesep}
\centering{}$\mathbf{\Lambda}, \mathbf{\Lambda_{\text{S}}}$ & diag. eigenvalue matrix, diag. eigenvalue submatrix
\tabularnewline[\doublerulesep]
\centering{}$\widetilde{\boldsymbol{\phi}}_\text{ROI}, \widetilde{\boldsymbol{\phi}}_\text{test}$, and $\widehat{\boldsymbol{\phi}}_\text{ROI}, \widehat{\boldsymbol{\phi}}_\text{test}$ & PCA-projected descriptor  of a query ROI\ and a test image (resp.), and whitening-and-normalization based product quantization (WNPQ)\ descriptor of the same
\tabularnewline[\doublerulesep]
\noalign{\vskip\doublerulesep}
\centering{}$B,B' $ & number of bits for quantized descriptor \&\ constituent block 
\tabularnewline[\doublerulesep]
\noalign{\vskip\doublerulesep}
\centering{}$Z,Z' $ & number of quantization centroids per descriptor and descriptor\ block\tabularnewline[\doublerulesep]
\noalign{\vskip\doublerulesep}
\centering{}$S_\text{des1,des2}$ & similarity score between descriptor ``des1'' and ``des2''\tabularnewline[\doublerulesep]
\noalign{\vskip\doublerulesep}
\centering{}$M$ & number of  quantization subspaces (blocks) for Product Quantization (PQ)
\tabularnewline[\doublerulesep]
\noalign{\vskip\doublerulesep}
\centering{}$\mathbf{C}_1, \dots \mathbf{C}_M$ & PQ\ codebook per  quantization\ block $m$, $1 \leq m \leq M$
\tabularnewline[\doublerulesep]
\noalign{\vskip\doublerulesep}
\centering{}$L$ & number of levels (scales) used for Voronoi-based encoding (VE)
\tabularnewline[\doublerulesep]
\centering{}$V_1,...,V_L$  & number of Voronoi cells per level $l$, $1\leq l \leq L$ 
\tabularnewline[\doublerulesep]
\noalign{\vskip\doublerulesep}
\centering{}$ V_\text{tot}$  & number of Voronoi cells in VE 
\tabularnewline[\doublerulesep]
\noalign{\vskip\doublerulesep}
\centering{}$l_\text{ph1}$ & level that Phase 1 exits in Fast-VE adaptive search, $0 \leq l_\text{ph1} < L$
\tabularnewline[\doublerulesep]
\noalign{\vskip\doublerulesep}
\centering{}$S^*_{0}, \dots ,S^*_{l_{\text{ph1}}}$ & similarity score for cell with maximum similarity to the query per level $l$, $0 \leq l \leq l_\text{ph1}$
\tabularnewline[\doublerulesep]
\noalign{\vskip\doublerulesep}
\centering{}$v_{0}, \dots ,v_{l_{\text{ph1}}}$ & difference between number of interest points in query and cell corresponding to $S^*_{l}$, $0 \leq l \leq l_\text{ph1}$
\tabularnewline[\doublerulesep]
\noalign{\vskip\doublerulesep}
\centering{}$\widehat w_{0}, \dots ,\widehat w_{l_{\text{ph1}}}$ & L1-normalized Gaussian weighting per level $l$ for Fast-VE, $0 \leq l \leq l_\text{ph1}$\tabularnewline
\noalign{\vskip\doublerulesep}
\centering{}$V_\text{F}$ & number of cells accessed in Fast-VE
\tabularnewline[\doublerulesep]
\noalign{\vskip\doublerulesep}
\centering{}$\mathbf{\Sigma}_1, \dots \mathbf{\Sigma}_M$ & covariance matrix per descriptor block $m$, $1 \leq m \leq M$ 
\tabularnewline[\doublerulesep]
\noalign{\vskip\doublerulesep}
\hline 
\noalign{\vskip\doublerulesep}
\end{tabular}
\end{table}

\subsection{Multi-VLAD}
For ROI-based retrieval, VLAD and the similarity measure of \eqref{similarity} will produce suboptimal results for small ROI, because information encoded from the remaining parts of the dataset image will distort the similarity scoring  \cite{arandjelovic2013all}. 

Lazebnik \textit{et al.} \cite{lazebnik2006beyond}  introduced the concept of spatially partitioning an image into a rectangular grid over multiple scales and encoding per block, as a  method of incorporating spatial information; this has found application in both image classification \cite{lazebnik2006beyond,mantziou2013large} and retrieval \cite{zhou2014spatial}.  Similarly, the recently-proposed Multi-VLAD\ descriptor \cite{arandjelovic2013all} attempts to improve VLAD performance for small ROI by spatially partitioning the dataset images into a rectangular grid over three scales and computing a VLAD descriptor per block.     

At the finest scale (level 2), nine VLADs are encoded over a 3\(\times\)3 rectangular grid.  At medium scale (level 1),  four VLADs are encoded over a 2\(\times\)2 grid, where each block is composed of 2\(\times\)2 blocks from the finest scale.  Finally, a single VLAD is encoded over the whole dataset image (level 0).  At each scale, Multi-VLAD excludes featureless regions near image borders by adjusting the grid boundary. Moreover, each VLAD\ is PCA projected and truncated to a 128-dimensional vector.  The similarity  is thus computed between the VLAD encoded over the query ROI and each of the 14 VLAD\ descriptors via \eqref{similarity} and the dataset image is assigned a similarity score to the ROI\ equal to the maximum similarity over its constituent VLADs. \
 
 For ROI queries occupying about 11\% of image real estate, the Multi-VLAD descriptor has been shown to outperform the single (128 \(\times\) 14)-D VLAD\ (computed over the whole image) in terms of mAP.  However, Multi-VLAD\ achieves 20\% lower mAP than the (128 \(times\) 14)-D VLAD when queries occupy a sizeable proportion of the image  \cite{arandjelovic2013all}. In addition, it incurs a 14-fold penalty in storage and matching complexity in comparison to the baseline 128-D VLAD. 
   
\subsection{Deep Convolutional Neural Networks for Retrieval}\label{ConvNet}
Deep CNNs  are feed-forward neural networks comprising multiple layers, and  typically   trained for classification on large \textit{a-priori} labelled datasets, such as ImageNet \cite{deng2009imagenet}.  It   has recently been shown that extracted features from intermediate layers are transferable to other visual recognition tasks, including image retrieval \cite{babenko2014neural, azizpour2014factors, yosinski2014transferable}.  While descriptors derived from these extracted features have been shown to match or outperform ``shallow'' learned methods, such as VLAD, they suffer due to lack of geometric invariance \cite{gong2014multi, chandrasekhar2015practical}. Recent works \cite{gong2014multi,razavian2014cnn,paulin2016convolutional,azizpour2014factors} have proposed patch-based methods for overcoming the lack of geometric invariance  of the descriptor in instance retrieval. Our proposal is more inline with grid based spatial search methods of Carlsson \textit{et al.} \cite{azizpour2014factors, razavian2014cnn},  as we are not explicitly computing a  global descriptor over extracted features from  multiple patches like CNN+VLAD \cite{gong2014multi} and CKN-mix \cite{paulin2016convolutional} (which require additional computational pre-processing, e.g., for learning encoding centers).  Therefore, we compare performance against a generic grid-based spatial search, which we refer to  as Multi-CNN. 
 
A Multi-CNN descriptor can be devised analogously to Multi-VLAD, i.e., by dividing the image at level $l$ into an $(l+1) \times (l+1)$ grid and computing the similarity score between two images as the global maximum inner product over all partitions. For the general case of $L$ levels, the total number of partitions (incl. the whole image as level 0) is:

\begin{equation}
P_\text{tot}=\sum^{L-1}_{l=0}(l+1)^2=\frac{(L+1)(L+2)(2L+3)}{6}
\label{MultiTot}
\end{equation}

In this paper we only consider networks pre-trained on ImageNet \cite{deng2009imagenet}. While fine-tuning on a tailored dataset may increase performance \cite{babenko2014neural},  this   requires additional training and diverts away from the generality of the CNN features extracted from a large and diverse dataset.      

\subsection{Product Quantization}\label{PQ_VLAD} 
\label{bground_PQ}

In order to further reduce the search complexity and the required memory footprint when handling large  datasets,  $D$-dimensional\ vectors are typically quantized  to produce  compact $B$-bit representations \cite{jegou2011product}. 

Consider a $D$-dimensional\ query vector  $\widetilde{\boldsymbol{\phi}} \in \mathbb{R}^D$. A global K-means clustering approach can be used to map $\widetilde{\boldsymbol{\phi}}$ to vector $q(\widetilde{\boldsymbol{\phi}})$ in  codebook  $\mathbf{C} = \left\{ \boldsymbol{c}_i, 1 \leq i \leq Z\right\}$.
 For quantizer $q$ with $Z$ centroids, the total number of bits used to encode $\boldsymbol{\widetilde{\boldsymbol{\phi}}}$ is $B=\log_2(Z)$.     However, for a 64-bit encoding of a 128-D \ query vector (0.5 bits per dimension), $Z = 2^{64}$ centroids must be learned using K-means, which is clearly infeasible. The learning and storage requirements of this quantization problem can be reduced either  via traditional approximate nearest neighbor (ANN) algorithms  \cite{datar2004locality,weiss2009spectral,torralba2008small, xu2011complementary}, or more recent advances \cite{li2015two,song2015top,ji2014query}. In this paper, we focus on an efficient method for ANN\ search,  named  product quantization (PQ) \cite{jegou2011product}, which uses multiple subquantizers rather than a single global quantizer \cite{jegou2011product}.
PQ
considers each unquantized vector $\widetilde{\boldsymbol{\phi}}$ as the concatenation of $M$ subvectors, $\widetilde{\boldsymbol{\phi}} = \left[\widetilde{\boldsymbol{\phi}}_1,\widetilde{\boldsymbol{\phi}}_2,\dots,\widetilde{\boldsymbol{\phi}}_M\right]$, each with equal dimension $D' = D/M$. Each subvector $\tilde{\boldsymbol{\phi}}_m$  is encoded from its own subcodebook $\mathbf{C}_m = \left\{ \boldsymbol{c}_{m,i}, 1 \leq i \leq Z'\right\}$, learned using K-means and considered to be of size $Z'=Z^{\frac{1}{M}}$ for all $m$, $1 \leq m \leq M$.   As such, the new codebook $\mathbf{C}$ is the Cartesian product of the subcodebooks, with total size $Z = (Z')^M$:
\begin{equation}
\mathbf{C} = \mathbf{C}_1 \times \dots \times\mathbf{C}_M.\label{codebook_C}
\end{equation}Crucially, via this vector partitioning approach, the learning complexity and storage requirement is reduced to $\mathcal{O}\left( MZ'D' \right)=\mathcal{O}\left( Z^{\frac{1}{M}}D \right)$.  The total number of bits used to encode each $\widetilde{\boldsymbol{\phi}}\in \mathbb{R}^D$ is now given by $B  = M \times B'$, where $B'$ is the number of bits used
to encode each subvector $\boldsymbol{x}_m$, i.e., $B' = \log_2(Z')=\frac{B}{M}$.

Previous work proposed PQ with asymmetric distance computation (ADC)   \cite{jegou2010aggregating, jegou2012aggregating}, which only encodes the vectors of the test dataset, and PQ with symmetric distance computation (SDC)  \cite{jegou2011product}, where both query and test vectors are quantized.  By not encoding the query vectors, ADC reduces the overall quantization distortion, thus enhancing the discriminatory power of the system. On the other hand, in SDC the distances between any two subcodewords in the $m$-th subspace are pre-computed and stored in a $Z'\times Z'$ lookup table, thus enabling efficient ANN search by simple lookup table accesses.   Experimental results   \cite{jegou2010aggregating, jegou2012aggregating}, have shown that ADC and SDC variants of PQ-based VLAD achieve comparable retrieval performance to unquantized VLAD representations with four to ten-fold reduction in storage and search complexity.  \

\section{Proposed Voronoi-based Encoding  and its Fast Online Implementation}
\label{sec:VLAD}

The Voronoi-based\ encoding proposed in Subsection \ref{sec:VVLAD_Encoding} constitutes the offline component of our system. Subsection \ref{Fast-VVLAD}  describes the proposed acceleration
for online Voronoi-based ROI query search and possibilities for memory compaction to reduce storage requirements.    
\subsection{Voronoi-based Encoding and Compact Descriptors}\label{sec:VVLAD_Encoding}
 Instead of spatially partitioning the images into a rectangular grid, we propose to partition the image into Voronoi cells over $L$  levels (scales), using hierarchical spatial K-means clustering.  The key intuition is that objects that may constitute ROI\ queries tend to appear as clusters of salient points, potentially interspersed with featureless regions in the image.  Therefore, a ROI-oriented partitioning must attempt to adaptively isolate these spatial clusters at multiple levels.

Initially, the entire image is encoded; this comprises level 0 of the Voronoi-based encoding. For level 1,  a spatial K-means is  computed over the interest point locations in the whole image,  which effectively partitions the image into \(V_1 \) Voronoi  cells.  Next, for level 2, a spatial K-means is computed over the interest point locations within each level-1 Voronoi cell, thus partitioning each cell into \(V_2\) constituent cells.  In general, for level \(l\), $1 \leq l < L$, each of the \(V_{l-1}\) cells of the previous level is partitioned into \(V_l\) cells, with $V_0\triangleq1$. A base descriptor, whether this be VLAD or aggregated deep CNN features, is encoded over  each cell following the description of Section 2.1, giving a total of 
\begin{equation} 
V_\text{tot}=1+ \sum^{L-1}_{l=1}\prod^{l}_{m=1} V_m 
\end{equation}
encodings per image.  When PCA-projecting each cell descriptor, we aggregate each level \(l\) into a single matrix \(\mathbf{\Phi}_{l}\).    

\begin{figure}[ht!]
\centerline{\includegraphics[width = 90mm, height = 22.5mm, scale = 0.1]{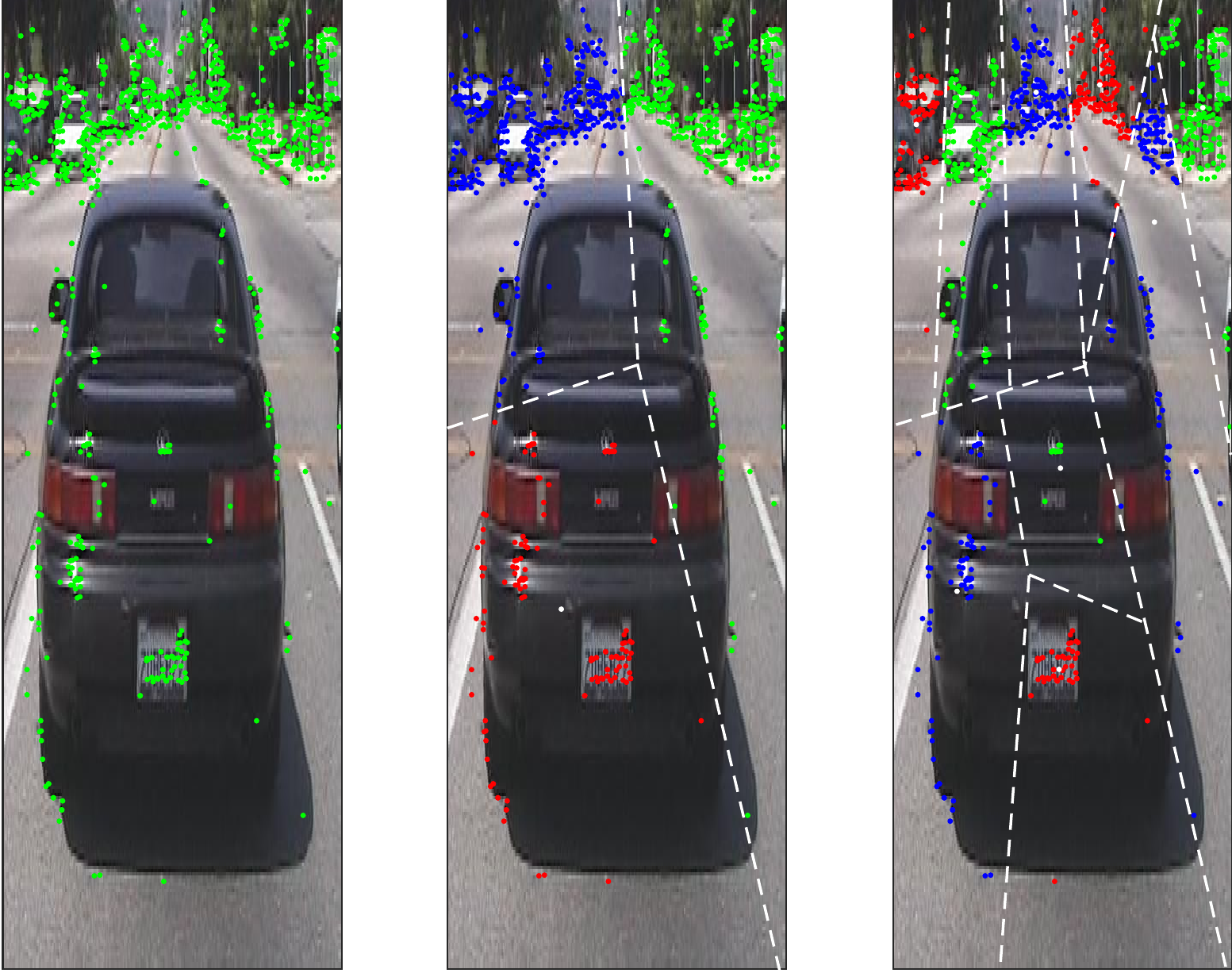}} 
\caption{Three-level Voronoi partitioning for an image from Caltech Cars dataset.  For illustration purposes, SIFT descriptors are color-differentiated for each cell.\label{overflow}}
\label{fig:car_voronoi}
\end{figure}

 A three-level Voronoi partitioning for an image from the Caltech Cars image dataset with $V_1=V_2=3$ is illustrated in Fig. \ref{fig:car_voronoi}. 
The detected points are shown in color in the left image of Fig. \ref{fig:car_voronoi}, and the level-1 and level-2 Voronoi cells are superimposed  with dashed lines on the middle and right image (resp.), with their corresponding descriptors appearing with different colors.  

In essence, there are two variables to consider when implementing a  Voronoi-based encoding; the number of levels $L$ and the number of Voronoi cells $V_{l}$, $1\leq l<L$, to encode.  For the purposes of this paper, we will consider $V_{l}$ to be constant for all levels $l$. In addition, it is worth noting that for the Voronoi-based encoding, we construct a single  PCA\  projection matrix using the entire images of the   training set. This is because we found that there is very little gain in retrieval performance when learning separate PCA\ projection matrices for each Voronoi partition level, mostly due to sufficient variability in scale of ROI in the training  images alone. Finally, given we are dealing with PCA on high dimensional data, for the case where the unprojected cell descriptor dimension, $U$, is greater than the training size $Y$, we use the manipulation described by Bishop \cite{bishop2006pattern}. In essence, we define the covariance matrix for the $U \times  Y$ training descriptor matrix $\mathbf{\Phi}_\text{T}$ as $\mathbf{\Phi}_\text{T}^T\mathbf{\Phi}_\text{T}$, which provides for a lower dimensional $Y \times Y$ matrix to work with.  Following singular value decomposition, we then rotate the derived projection matrix, $\mathbf{R_\text{ROT}}$, into the original covariance data space to obtain the $U \times D$ projection matrix, $\mathbf{R}$, using the  equivalence: 

\begin{equation}
\mathbf{R} = \mathbf{\Phi}_\text{T}\mathbf{R_{\text{ROT}}}\mathbf{\Lambda}^{-1},
\label{eq:R}
\end{equation}
where $\mathbf{\Lambda} = \text{diag}(\lambda_{1}^{-\frac{1}{2}},\lambda_{2}^{-\frac{1}{2}},\dots,\lambda_{Y}^{-\frac{1}{2}})$ is the diagonal matrix of eigenvalues of $\mathbf{\Phi}_\text{T}^T\mathbf{\Phi}_\text{T}$. 

We conclude this subsection by summarizing the  VLAD and deep CNN descriptors utilized for each Voronoi partition.

\subsubsection{Voronoi-based VLAD (VVLAD)}

We require a detector that is robust to scale and viewpoint changes, while also detecting enough points in salient regions to allow for reliable partitioning. Therefore, for VVLAD, we use the Hessian Affine detector \cite{mikolajczyk2005comparison,mikolajczyk2002affine}, which is based on the multi-scale determinant of the Hessian matrix (computed locally), and detects affine covariant regions.  SIFT descriptors are  produced based on the detected points. It is worth noting that: \textit{(i)} salient point detection is an implicit step in each VLAD\ computation and not additional processing; \textit{(ii)} unlike Multi-VLAD, there is no need to preprocess the image and exclude featureless regions. As shown in the example of Fig. \ref{fig:car_voronoi}, smaller Voronoi cells are adaptively formed around  regions of tight clusters of detected points.

\subsubsection{Voronoi-based Deep CNN (VDCNN)} 

In this case, the salient point detection constitutes additional pre-processing.  Nevertheless, this can be achieved efficiently by using the FAST corner detector \cite{rosten2006machine}, which classifies a pixel as a corner based on its relative intensity to a set of contiguous pixels.  As for the case of VVLAD, the image is partitioned into Voronoi cells based on the location of detected points. Since the deep CNN must take a rectangular input image segment, we compute a bounding box over the constituent points of each cell, and then resize and subtract an average image, as per convention, before feeding into the pre-trained deep CNN. Given that the cells are treated independently, the feed-through can be done in parallel, using multiple copies of the network. In terms of the deep CNN descriptor specifics, we use the CNN-S architecture \cite{chatfield2014return} pretrained on ILSVRC-2012 with batch normalization \cite{ioffe2015batch}.  The network is sufficiently deep to provide a rich semantic representation of the image/image partitions without overfitting to the classification task.   The conventional approach to generating a feature descriptor from the network is  to simply extract one of the fully-connected layers \cite{babenko2014neural, krizhevsky2012imagenet, razavian2014cnn}.  Instead, we extract the last max-pooling layer (Layer 13) of the network, which precedes the fully-connected network and should be less tuned to the classification task. From this layer, we generate a 512-D feature descriptor by  averaging  the CNN activations over the spatial dimensions. We can also (optionally) apply PCA-projection and truncation to achieve further compaction to 128 dimensions.  
\

\subsection{Fast Online Implementation: Adaptive Search and  Image Similarity Score}\label{Fast-VVLAD}
Conventionally, we could assign an image score as the global maximum similarity to a query over cells, using \eqref{similarity} for each cell. However, the proposed Voronoi partitioning essentially gives us a  tree of spatial Voronoi cells where, for $L$ levels, $\prod^{L-1}_{l=1} V_l$ ``leaf'' Voronoi cells exist at the bottom of the tree. Given that there is inherent mutual information between a cell and its constituent cells, rather than accessing data for all levels and measuring  similarity over all \(V_{\text{tot}}\) cells of the tree indiscriminately, we can design an adaptive search with top-to-bottom tree pruning to find the most relevant Voronoi cells to the query.  This  reduces the overall execution time and memory accesses when performing a retrieval task, which makes our proposal applicable to very large image databases  that would contain millions of images.
The top-to-bottom search is carried out in two phases.   

\textbf{Phase-1:} Considering the cell of level \(l-1\) with maximum similarity to the query [measured via \eqref{similarity}], in Phase-1 of the search, we assume that either this cell or a constituent cell within it (at level $l$) will attain high similarity to the query. If the cell of level \(l-1\) is found to attain the highest similarity to the query, we terminate the search for that image at level $l-1$ and proceed to Phase-2. On the other hand, if we find that a constituent cell of level $l$ attains the maximum similarity, we repeat Phase-1 for that  cell and its constituent cells at the next level ($l+1$), until we reach the bottom of the tree, in which case we move to Phase-2.

\textbf{Phase-2:} Let us denote the maximum similarity found by Phase-1 for each level \(l\) as \(S^{*}_{l}\) and assume that Phase-1 exited at level $l_\text{ph1}$, $0 \leq l_\text{ph1} < L$.  Rather than assigning \(S^{*}_{l_{\text{ph1}}}\) as the similarity score between the ROI query and the test image $I$ in the dataset, we perform a weighted sum over all  $S^*_{0}, \dots ,S^*_{l_{\text{ph1}}}$.  To this end, we first compute the  difference \(v_{l}\), \(0\leq l \leq l_\text{ph1}\), between the number of interest points in the query and the number of interest points in the image dataset cell corresponding to \(S^*_{l}\).  This difference is subsequently used within a scaled inverse function.  The weight for  \(S^{*}_{l}\)   (\(0\leq l \leq l_\text{ph1}\)) is thus defined as: 
\begin{equation}
w_l = \frac{C}{\max{(\left\vert v_l \right\vert, 1)} }.
\end{equation} 
where $C$ controls the order (set as  the modal order of magnitude over all $v_l$).  The weight vector over all levels is  \(L_1 \)-normalized so that the image score can be ranked independently of the level $ l_\text{ph1}$ at which Phase-1 terminated. Denoting the $L_1$-normalized weight  as \(\hat w_l\), the proposed similarity score between a ROI query and dataset image $I$ after Phase-2 is:  
\begin{equation}
S_{\text{ROI},I} = \sum^{l_{\text{ph1}}}_{l=0}\hat w_{l} S^*_{l}.\label{S_ROI_I}\end{equation}
For example, for a three-level partition, if a query object  is small relative to the image size, we expect that the total number of interest points over the query would be comparable to that of a level-2 cell. Hence, the level-2 maximum dot product \(S^{*}_{2}\) should receive the largest weighting \(\hat w_{2}\) when computing the   similarity score.   This is expected to be a more robust similarity scoring than just taking a global maximum over all  $S^*_{0}, \dots ,S^*_{l_{\text{ph1}}}$ (as in Multi-VLAD) as the similarity score, since we account for relevant information from all levels. 

\textbf{Summary:} We term this two-phase search coupled with the Voronoi-partitioning  as  \textit{Fast Voronoi-based encoding} (Fast-VE), because it reduces the expected number of cells that are accessed at runtime.  The upper bound for the  matching complexity\ is now: 
\begin{equation}
V_\text{F}= 1+\sum^{\text{min}\left\{l_{\text{ph1}}+1,L-1\right\}}_{l=1}V_l
\label{V_F}
\end{equation} 
inner products per image instead of  the \(V_\text{tot}\) inner products required using a global maximum similarity measure that considers all cells. Due to the weights of \eqref{S_ROI_I}, per image $I$, along with the  Voronoi-based encoding we also store the  number of interest points per cell, comprising $V_\text{tot}$ additional values. 

It is worth noting that  further storage compaction of the Fast-VE is feasible using level projection. Via level projection, we can adhere to memory constraints  of a practical deployment for very-large image datasets by only storing  the PCA projected cell descriptors for the last level, $L-1$ and computing\  the cell descriptors for levels $0,...,L-2$ at runtime by aggregating smaller-cell descriptors. Given that such storage compaction is of secondary importance in the overall unquantized and quantized Fast-VE design, we include its details as supplementary information in Appendix \ref{App A}.

\section{Product Quantization for\ Efficient VE Search}\label{sec:PQ}

Given that quantized descriptor representations offer significantly-higher compaction than unquantized ones, we extend VE\ and Fast-VE to quantized representations via a specially-designed product quantization framework.

\subsection{ Product Quantization based on Symmetric Distance Computation for Voronoi-based Encoding }

We consider  PQ based on SDC for the proposed VE approach\footnote{refer to Section \ref{PQ_VLAD} for nomenclature and symbol definitions}, where both the query vector $\boldsymbol{y}$ and test dataset vector $\boldsymbol{t}$ are quantized \cite{jegou2011product}.  We opted for SDC-based rather than ADC-based PQ because ADC-based methods require the precomputation and storage of distances between VE query and test vectors, which is not feasible in a large-scale image retrieval system where potentially any image could form a query.  

In SDC-based PQ, the  nearest neighbour to $\boldsymbol{y}$ can be approximated by optimizing the distance function $d(q(\boldsymbol{y}),q(\boldsymbol{t}))$. The distance function is typically the squared Euclidean distance \cite{jegou2011product}: \begin{eqnarray}
d(q(\boldsymbol{y}),q(\boldsymbol{t})) & = & \left\Vert q(\boldsymbol{y})-q(\boldsymbol{t}) \right\Vert^2 \nonumber \\ & = &  \sum_{m = 1,\dots,M}d(q_m(\boldsymbol{y}_m),q_m(\boldsymbol{t}_m))  \\  & = &  \sum_{m = 1,\dots,M}\left\Vert q_m(\boldsymbol{y}_m)-q_m(\boldsymbol{t}_m) \right\Vert^2,\nonumber 
\end{eqnarray}
with $M$ the number of subquantizer blocks of \eqref{codebook_C}. 

The key intuition behind the modified PQ for VE is  to treat the constituent Voronoi cells as images and apply PQ on each query and test  cell. 
A single PQ\ codebook $\mathbf{C}$ is learned using K-means clustering on a training set.  Each cell descriptor from the test dataset is thus  considered  as a concatenation of $M$ subvectors  $\widetilde{\boldsymbol{\phi}}=\left[\widetilde{\boldsymbol{\phi}}_1, \dots,\widetilde{\boldsymbol{\phi}}_M\right]$ of $D'=D/M$ elements each, with each subvector being encoded from its corresponding subcodebook $\mathbf{C}_m$. In this way, there is also no dependency on the level $l$, as we quantize the cell descriptors from a single PQ codebook. All possible distance values between the $i$th and $j$th subcodebook vectors in the $m$-th subspace, $d(\boldsymbol{c}_{m,i},\boldsymbol{c}_{m,j})$, are pre-computed and stored in a $Z'\times Z'$ lookup table, thus enabling efficient ANN search by simple lookup table accesses. 

As the subspaces are orthogonal, we $L_2$-normalize the product quantization of each  cell descriptor's subquantizer block $m$, $1\leq m \leq M$, by normalizing the columns of the PQ subcodebooks  $\mathbf{C}_m$  individually before computing and storing the distance values.  For the $i$th subcodebook vector in the $m$-th subspace, the normalization term is given by $\sqrt{M}\left\Vert \boldsymbol{c}_{m,i} \right\Vert$.  As such, the distance value to be stored between the $i$th and $j$th subcodebook vectors is

\begin{equation}
d(\boldsymbol{c}_{m,i},\boldsymbol{c}_{m,j}) =  {\frac{\left\langle \boldsymbol{c}_{m,i}, \boldsymbol{c}_{m,j} \right\rangle}{M\left\Vert \boldsymbol{c}_{m,i} \right\Vert\left\Vert \boldsymbol{c}_{m,j} \right\Vert}}.
\end{equation}

 Quantizing from the normalized  PQ subcodebooks,  the distance function between a  subspace normalized and quantized query\ cell descriptor $q(\boldsymbol{y})$ and a subspace normalized and quantized test cell descriptor $q(\boldsymbol{t})$ is  now simply 

\begin{equation}
d(q(\boldsymbol{y}),q(\boldsymbol{t}))  =  \sum_{m = 1,\dots,M}{\left\langle q_m(\boldsymbol{y}_m), q_m(\boldsymbol{t}_m) \right\rangle},
\label{eq:VVLAD_dist_norm}
\end{equation}
which is analogous to the squared Euclidean distance of normalized vectors.
Importantly, this bounds the similarity score  between -1 and 1, which facilitates performance comparisons. 

\begin{figure}[ht!]
\centerline{\includegraphics[scale = 0.25]{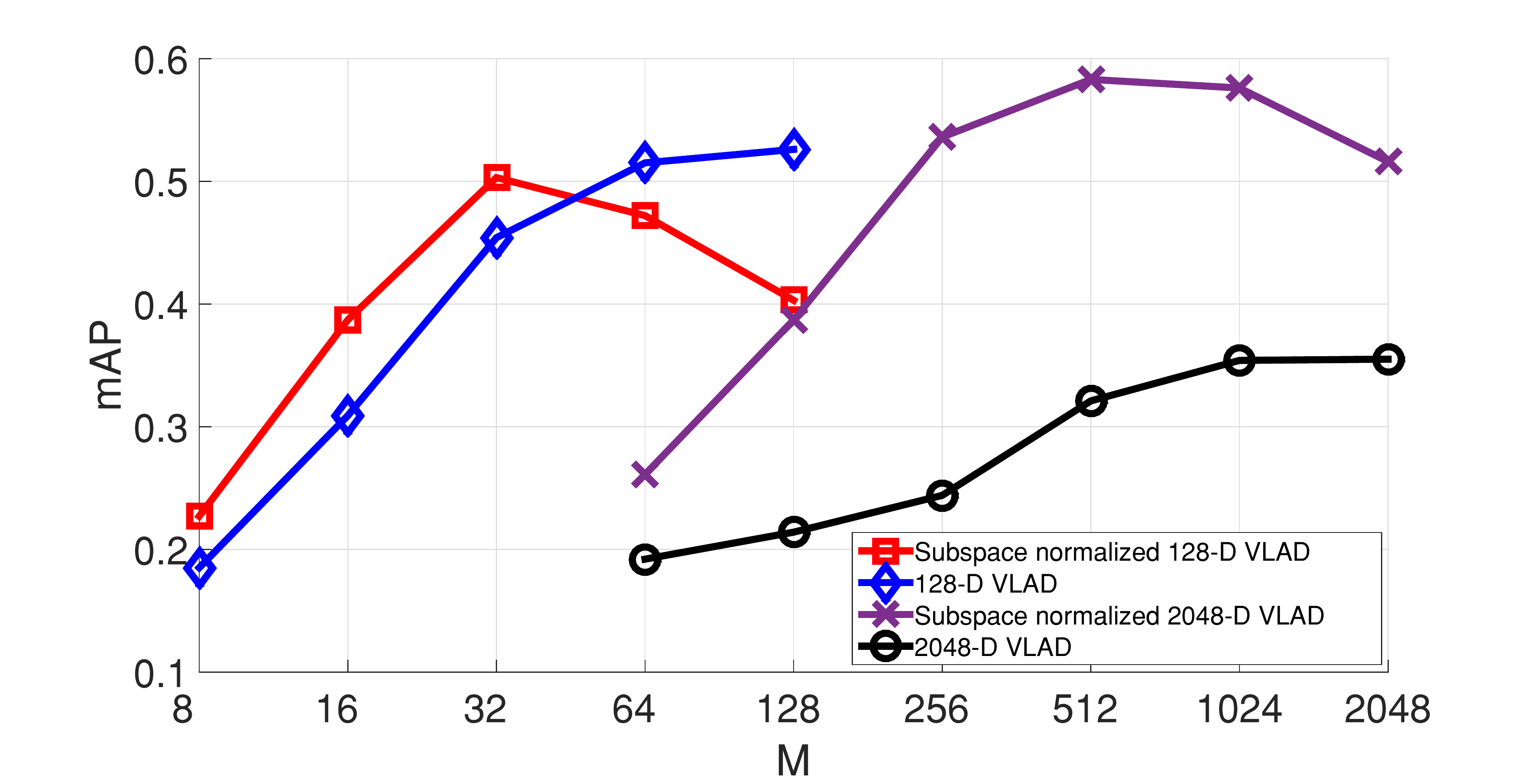}}
\caption{Plot of mean average precision (mAP) with varying number of PQ blocks, $M$, for PQ VLAD descriptors on the Holidays dataset. \textbf{}\textbf{}}
\label{fig:block norm}
\end{figure}

Fig. \ref{fig:block norm} illustrates two indicative examples of SDC-based PQ on the $D=128$ and $D=2048$ dimensional VLAD, both with and without subspace normalization.  The retrieval performance is measured in terms of mean average precision (mAP) on the Holidays dataset \cite{jegou2008hamming}, using whole-image queries.  It is evident from the results of the figure that, for given dimension $D'=D/M$,  subspace normalization actually improves retrieval performance, effectively peaking close to $M = D/4$. At this block size, the subspace dimensionality is sufficient such that each subspace is optimally regularized.  In addition, we observe that the performance margin between the VLAD descriptor and its subspace normalized counterpart increases significantly with dimension $D$.

Essentially, we want the block size to be large enough that we  encode over a sufficient number of bits; however, beyond a certain block size, we end up normalizing over too few dimensions. In this regard, it is interesting to consider the limit case, where $M=D,$ i.e.,  $D'=1$. There,  subspace normalization results in storing just the sign per cell descriptor.  In this extreme case, the similarity between cells can be computed very efficiently by using the Hamming distance, i.e., without accessing any lookup tables.

Concerning storage requirements, assuming that the components of the unquantized cell representation are kept as 32-bit floating-point numbers, their offline storage requirement is $D \times 32$ bits per test image. On the other hand, our product-quantized cell descriptor\ requires   $V_\text{tot} \times B$ bits per test image, which is independent of the\ dimension, $D$. In addition, for the entire test dataset, the total storage requirement for the quantization lookup tables is $Z' \times Z' \times M$. As the test dataset grows in size,  this value becomes negligible in comparison to the storage requirement for the product-quantized descriptors.

Finally, with regards to the search complexity, the inner products have been replaced by read accesses to the look-up tables.  As such, the  product quantized Fast-VE\  now has an upper bound on complexity of $M \times V_\text{F}$ reads, which is independent of the descriptor\  dimension per cell.

\subsection{Optimal Bit Allocation via Whitening and Subspace Normalization}

Given the presence of multiple cells in the Voronoi-based encoding, it is important to derive an appropriate bit allocation strategy that minimizes the quantization distortion. \\
\\
\textbf{Assumption 1.} \textit{We consider successive samples of each subspace-normalized VE\ component (dimension) $\widetilde{\phi}_i$ ($1\leq i \leq D/M$) to be modelled by independent, normally-distributed, random variables, with corresponding variance $\sigma_i$.}

Under Assumption 1, the normalized random vectors  $\widetilde{\boldsymbol{\phi}}_m$    of all subspaces $m$, $1\leq m \leq M$, can then be represented by independent and identically-distributed multivariate Gaussians\footnote{The Gaussian assumption is necessary for some of the theoretical derivations, but is also proven to hold in practice \cite{brandt2010transform,ge2014optimized}.}, with corresponding diagonal covariance matrices $\mathbf{\Sigma}_m$. The rate-distortion function for independent, normally-distributed random variables \cite{cover2012elements} can be extended to the multivariate case in order to derive the optimal bit allocation strategy for VE. This leads to the following proposition.

\begin{prop}
Under Assumption 1, optimal bit allocation after subspace normalization in VE can be achieved by balancing the variances of the subspaces.
\label{prop1}
\end{prop}

\begin{IEEEproof}
See Appendix B.
\end{IEEEproof}

Indeed, recent work \cite{ge2014optimized} employs an optimized product quantization (OPQ) that  effectively leads to balanced subspace variances by assigning principal components to a subspace with the objective of balancing the product of eigenvalues per subspace. This corresponds to performing a permutation of the principal components to achieve balanced variances. Jegou \textit{et al.} \cite{jegou2012aggregating} propose balancing the component variance with a random orthogonal rotation, but this removes the decorrelation achieved by PCA. A different approach is proposed by Brandt \textit{et al.} \cite{brandt2010transform}: one can achieve a constant quantization distortion per subspace by varying the number of bits assigned to each principal component, at the cost of increased training and runtime complexity. Finally, Spyromitros-Xioufis \textit{et al.} \cite{spyromitros2014comprehensive} consider the effects of  applying a random orthogonal rotation on PCA-projected and whitened VLAD vectors prior to product quantization. However, whitening inherently balances the subspace variances by setting $\mathbf\Sigma_m$  to the identity matrix  for all $m$, which also preserves  decorrelation and  mitigates  descriptor bias from visual word/component co-occurrences \cite{jegou2012negative,chum2010unsupervised}. As such, we propose a  simple and effective solution for the bit allocation that adheres with the theoretical result of Proposition \ref{prop1}: we  use a  whitening approach after PCA (and prior to the product quantization), together with the subspace normalization described in the previous section (and shown to be beneficial by the tests of Fig. \ref{fig:block norm}). Specifically, per cell, we can express the relationship between  a projected descriptor $\widetilde{\boldsymbol{\phi}}_m$ and its whitened and normalized counterpart $\widehat{\boldsymbol{\phi}}_m$ in the $m$-th subspace as:

\begin{equation}
\widehat{\boldsymbol{\phi}}_m = \frac{\mathbf{\Lambda}_{\text{S}, m} \widetilde{\boldsymbol{\phi}}_m}{\left\Vert \mathbf{\Lambda}_{\text{S}, m} \widetilde{\boldsymbol{\phi}}_m \right\Vert},
\end{equation}
where $\mathbf{\Lambda}_{\text{S},m}=\text{diag}(\lambda_{a+1}^{-\frac{1}{2}},\lambda_{a+2}^{-\frac{1}{2}},\dots,\lambda_{a+D'}^{-\frac{1}{2}})$ is the diagonal subspace matrix of eigenvalues of the training-set covariance matrix, with $\lambda_i$ associated with the $i$-th largest eigenvector and $a=D'(m-1)$. 

The advantage of using whitening and normalization against previous approaches   is that there is no need for any additional pre-processing, such as learning a rotation matrix or variability in the bit allocation across the principal components.   We term our approach  \textit{whitening \&\ normalization based product quantization} (WNPQ).

\section{Experimental Evaluation}\label{sec:Experiments}

\subsection{Datasets} \label{evalDat}

\begin{figure}[ht!]
\centerline{\includegraphics[scale = 0.17]{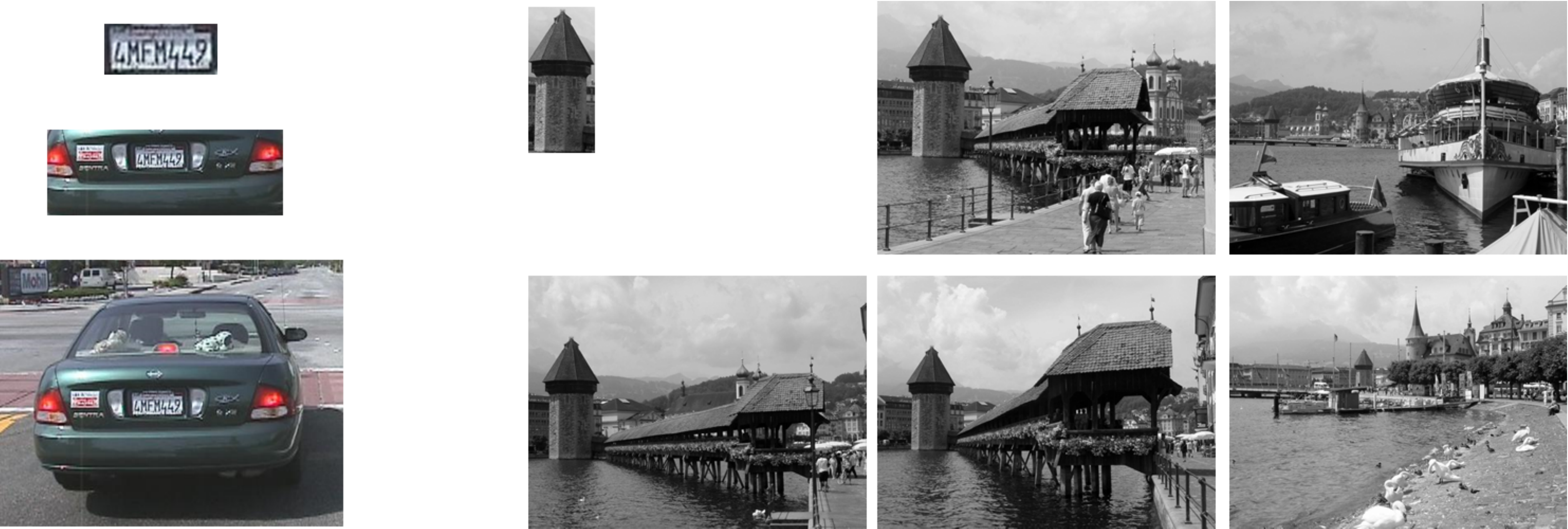}}
\caption{(Left) Example queries for the Caltech Cars dataset.  (Right) Example ROI query (top left) and matching image set for the Holidays dataset (remaining images).\label{overflow}}
\label{fig:queries}
\end{figure}

We measure performance on the Holidays and Caltech Cars (Rear) test image datasets. For both datasets, a set of predefined queries and hand-annotated ground truth is used. 
\\ \\\textbf{Caltech + Stanford Cars} \cite{fergus2003object, krause20133d}: 
 This test dataset consists of 1155 (360 $\times$) 240) photographs of cars taken from the rear.     Subsequently, we  test on a subset of 416 images from the Caltech Cars (Rear) dataset, from which we select 10 images  and perform three tests: \textit{(i)} we mimic a surveillance test by selecting only the license plates as ROI-queries; \textit{(ii)} we select as mid-scale ROI-queries a section of the car trunk, and \textit{(iii)} use the whole images as queries.   An example of the query subset is given in the left part of Fig. \ref{fig:queries}. For the license plate  test, we manually create ``good'' and ``junk'' ground-truth files over matching  images \cite{philbin2007object}; ``junk'' ground truth comprises any image in which the query (i.e., the license plate) is barely visible or  not distinguishable by the interest point detector. To provide a more rigorous and diversified test, we combine the Caltech Cars subset  with another independent set of 1000 distractor images from the Stanford Cars dataset \cite{krause20133d}, comprising  various car models and orientations, giving the Caltech + Stanford Cars dataset. \\ \\ \textbf{Holidays} \cite{jegou2008hamming}: The Holidays test dataset consists of 1491 images, mainly consisting of holiday photos.  There are 500 ``whole image" provided queries of a distinct scene or object.  In order to test on a  smaller scale, we also select  salient regions from a subset of 40 query images as ROI queries into our system. An example ROI query with its corresponding matching image set is shown in the right part of Fig. \ref{fig:queries}.

%
%

\subsection{Setup} \label{expDet}
Unless stated otherwise, all  vectors are whitened and re-normalized post-PCA.  The retrieval performance is measured by creating a ranked list and computing the mAP over all queries.   Matching complexity is defined as the number of multiply-accumulate (MAC) operations for unquantized descriptors, or the number of look-up table reads for quantized descriptors. Per descriptor, we report the matching complexity averaged over all tests and normalized to the baseline 128-D descriptor complexity, along with the descriptor storage size, in bytes. \\ \\\textbf{Caltech + Stanford Cars}: {Due to the specificity of the Caltech + Stanford Cars dataset, together with the lower ROI resolutions, using a deep CNN pre-trained on ImageNet is not a viable option. For example, ImageNet (ILSVRC2012 dataset \cite{ioffe2015batch}) does not contain any substantial number of images (and associated labels) corresponding to car license plates; therefore, the pre-trained deep CNN descriptor will not be suitable for such images. For these reasons, we have established that, in this dataset, the utilized deep CNN descriptor is outperformed by VLAD\ descriptor variants, particularly on license-plate queries. \    Thus, we use this dataset to test how the proposed  Voronoi-based encoding performs with the ``shallow'' learned VLAD descriptor of Subsection \ref{sec:VVLAD_Encoding}.1.} 

For the VLAD computation, we follow the design of Subsection \ref{VLADdesc}.  The PCA projection matrix, visual word centers and PQ codebook are learned on an  independent dataset of 2000 car images from the Stanford Cars dataset \cite{krause20133d}. For  Fast-VVLAD, we set:  \(K=64\),  \(L=3 \),  \(V_1=V_2=3\), with 128-D VLAD per cell and compile a ranked list from the relevant similarity score.
 For VLAD, we use: 128-D,  768-D and 1664-D sizes, in order to align the VLAD\ matching complexity with that of the Fast-VVLAD\ descriptor. We set $L = 3$  for the  Multi-VLAD descriptor, such that it is inline with our Voronoi partitioning.  This results in  a (128 \(\times\) 14)-D descriptor size per image  \cite{arandjelovic2013all}, as derived from \eqref{MultiTot}.

\textit{WNPQ Parameter Selection:}   
 We set $M = 32$  for the quantized 128-D VLADs. For the 768-D and 1664-D quantized VLADs,  we respectively set the  block size to $M =96$ and $M = 208$. For the quantized VVLAD, we set  $M = 32$  for all cell VLADs.    Finally, quantized Multi-VLAD also uses  $M = 32$. These settings for the block size were chosen to align the matching complexity of the quantized 1664-D VLAD\ with that of the quantized Fast-VVLAD, whilst providing the  768-D VLAD\ as a solution with mid-range complexity.    Notably, we fix $Z'=256$ for all experiments.  Higher values for $Z'$ increase the computational load for each block quantizer, whilst increasing the storage requirement of the look-up tables ($\mathbb{O}(Z'^2))$, which is an important detriment as these tables need to be sufficiently small to fit in cache memory \cite{jegou2011product}. \\ \\\textbf{Holidays}: The Holidays dataset provides a less controlled  test for our system.  The scenes in the Holidays dataset are better represented by a deep CNN architecture trained on ImageNet, particularly due to their high resolution.  {Similar to prior work \cite{babenko2015aggregating}, we have confirmed that deep CNNs substantially outperform VLAD descriptors for this dataset. Therefore, we use this dataset to test how the proposed Voronoi-based encoding performs with the deep CNN  descriptor of Subsection \ref{sec:VVLAD_Encoding}.2.} 

For the utilized CNN-S architecture \cite{chatfield2014return}, all images and  image partitions are resized to $224 \times 224$ and fed into the network after subtracting an average image.  The final feature descriptor is 512-D, which can then be normalized, PCA-projected to 128-D and whitened.  However, following a similar approach to the instance retrieval pipeline on the VLAD descriptor, we normalize, sign-square root and re-normalize the feature descriptor  prior to PCA and whitening, with the intention of minimising the burstiness of dimensions and thus adding to descriptor invariance \cite{jegou2012aggregating}. {It is worth noting that contrary to recent works \cite{babenko2015aggregating, radenovic2016cnn}, we do not manually rotate the images in the Holidays dataset as we do not deem this to be a fair representation of data `in the wild'.} 

\textit{Parameter selection for the Voronoi partitioning:} For the VE of all cases, the FAST corner detector \cite{rosten2006machine} is used, and we learn the PCA projection matrix and PQ codebook on a subset of 4000 images from the ILSVRC-2010 validation set.  For Fast-VDCNN, we set:  \(L=3 \),  \(V_1=V_2=3\), with a 128-D CNN feature descriptor per cell and compile a ranked list from the relevant similarity score. We compare this with a 128-D and the full unprojected 512-D CNN feature descriptors.  For Multi-CNN, we use the same grid partitioning as Multi-VLAD, with $L = 3$, thus producing a $(128 \times 14)$-D size per image.

\textit{WNPQ\ Parameter Selection:} We set $M = 32$  for the quantized 128-D CNN feature descriptors. For the 512-D quantized CNN feature descriptor,  we  set the  block size to $M =128$. For the quantized Fast-VDCNN, we set  $M = 32$  for all cell descriptors.    Finally, quantized Multi-CNN also uses  $M = 32$. As with the VLAD descriptors, we fix $Z'=256$ for all experiments to keep the storage requirement for the lookup tables constant.

\subsection{Results with Unquantized Descriptors}
This section summarises performance using unquantized descriptors on the Caltech + Stanford Cars and Holidays dataset.
\\ \\\textbf{Caltech + Stanford Cars}:
Table \ref{tab:mAPCars} summarizes the retrieval performance of all unquantized VLAD methods on the Caltech + Stanford Cars dataset. The first observation is that the   Fast-VVLAD\ descriptor  offers competitive performance to the larger  1664-D VLAD, whilst decreasing the matching complexity by more than 50\%. \ In addition, Fast-VVLAD performs significantly better  on  license plate queries than both the 128-D VLAD and its 768-D VLAD\ complexity counterpart, yielding respective mAP gains of  over 200\% and 41\%. Fast-VVLAD  maintains consistently-good mAP even with the larger ROIs of car trunks and whole-image queries, and is only outperformed on whole-image queries  by VLAD by (up to) a 7\%  margin. Finally, Fast-VVLAD  maintains competitive performance   to Multi-VLAD on all query types, whilst offering lower dimensionality and matching complexity.
\\ \\\textbf{Holidays}:
  Table \ref{tab:mAPHol} summarises the retrieval performance for the 500 whole-image queries and 40 smaller ROI\ queries on the Holidays dataset. Interestingly, the Fast-VDCNN  remains competitive on whole image queries. This is attributed to the Fast-VDCNN\ similarity score of \eqref{S_ROI_I} that considers all partition levels, which  provides  robustness against false positives. The Fast-VDCNN  is  found to outperform  Multi-CNN for whole image queries and maintain very competitive performance on ROI queries, while offering more than 50\%\ reduction in the matching complexity\footnote{We have also validated that this saving translates to practical runtime saving: by adding a large distractor set (thereby scaling the dataset size to 150K images), we found that Fast-VDCNN based retrieval is 40\% faster than Multi-CNN retrieval, with execution time comparable to the baseline 512-D CNN feature descriptor.}. Fast-VDCNN was also found to substantially outperform the lower dimensional CNN feature descriptors\ for ROI\ queries (gains exceeding 50\% in mAP). Given that the utilized CNN-S descriptor derived from Layer 13 is limited to 512 dimensions  \cite{chatfield2014return}, we also benchmarked using the first fully connected layer (FC1), which allows for a large 4096-D feature descriptor. Nevertheless, the FC1\ descriptor performed significantly worse than our 512-D Layer 13 descriptors for both query ROI and whole images, scoring mAP of 28.3\% and 71.4\%, respectively. This serves as an additional validation for our choice for the utilized CNN layer.

\subsection{Results with Quantized Descriptors}

\begin{figure}[ht!]
\centering
\centerline{\includegraphics[scale = 0.15]{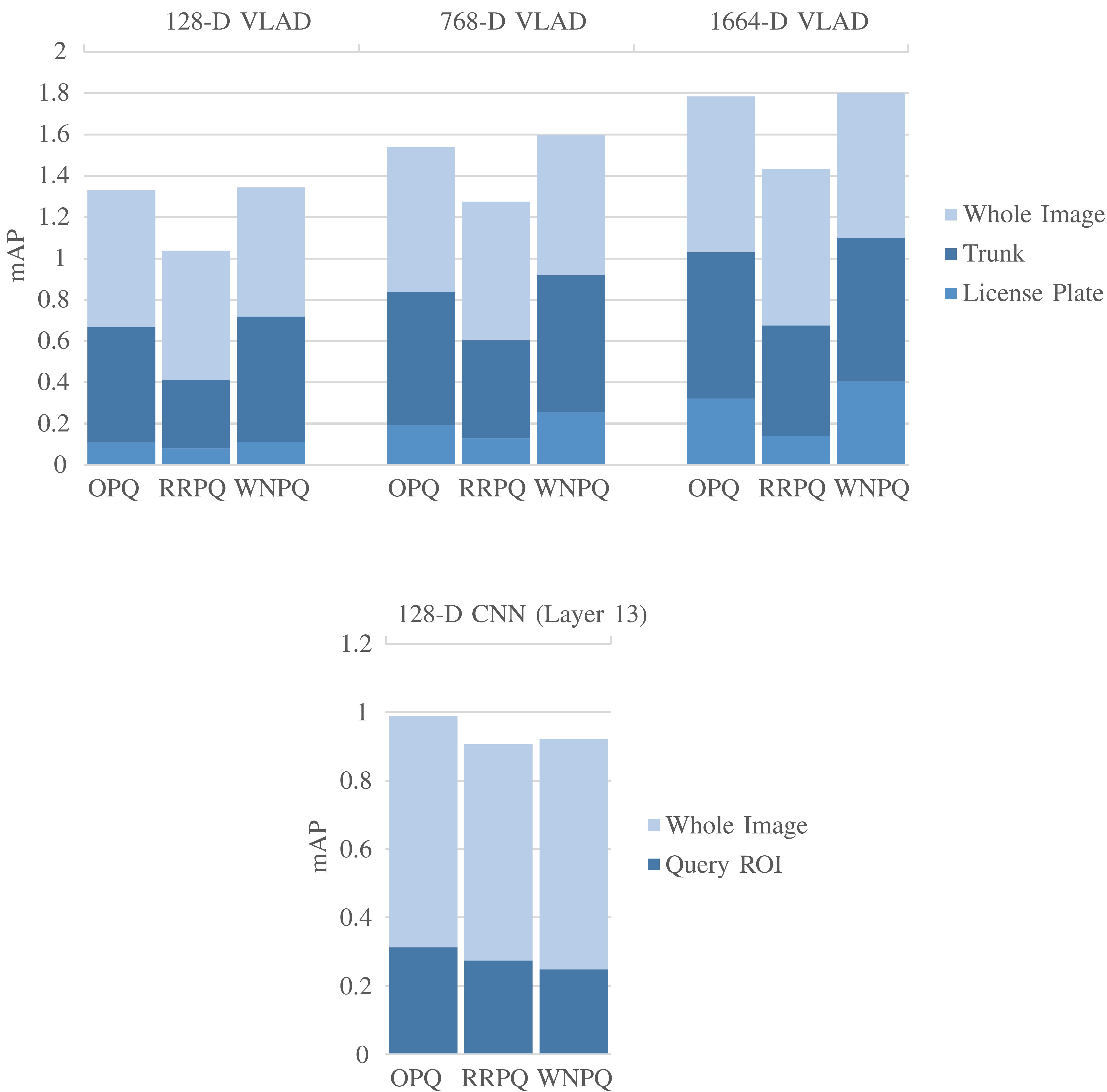}}
\caption{ Comparison of quantization methods. Top: Caltech + Stanford Cars dataset: mid blue = license plate, light blue = whole image, dark blue = trunk. Bottom: Holidays (128-D): dark blue = query ROI, light blue = whole image.} 
\label{PQ_cmp}
\end{figure}

We now consider performance when integrating  quantization into all approaches under consideration.  \\ \\\textbf{WNPQ against other quantization methods:}\ We first consider the performance of the proposed WNPQ method against other state-of-the-art methods,  namely the parametric optimized product quantization (OPQ) \cite{ge2014optimized}  and product quantization with a random rotation pre-processing (RRPQ)\cite{jegou2012aggregating}. As the OPQ and RRPQ descriptors are not normalized, we use the squared Euclidean distance metric for these methods and compare\ retrieval performance on both datasets.  The results of Fig. \ref{PQ_cmp} show that the proposed WNPQ method  outperforms RRPQ and, for the majority of the tests, also outperforms OPQ. Essentially, the WNPQ maintains its high retrieval performance when the dimensionality is increased from 128-D to the 768-D and 1664-D  VLAD descriptors. To ensure a fair comparison, and because the proposed WNPQ\ was shown to provide for the best overall performance, we use it to quantize all the descriptors under comparison. \\ 

\begin{table}
\caption{Complexity and mAP\ results for the Caltech + Stanford Cars image dataset \cite{fergus2003object, krause20133d}.}
\centering
\resizebox{\columnwidth}{!}{
\begin{tabular}{cc>{\centering}p{1.5cm}>{\centering}p{1cm}>{\centering}p{1cm}>{\centering}p{1cm}>{\centering}p{1cm}}
\hline\noalign{\smallskip}
 & $D_\text{tot}$  &  Matching Complexity & Descriptor Storage (bytes) & License\\ Plates & Trunk & Whole Image\tabularnewline
\noalign{\smallskip}
\hline
\noalign{\smallskip} 
VLAD \cite{jegou2012aggregating}  & 128 & 1 & 512 & 0.148 &0.669 &0.729\tabularnewline

& 768 & 6 & 3.07k & 0.348 & 0.739 & 0.780\tabularnewline

& 1664 & 13 & 6.66k &\textbf{0.512} &0.722 &\textbf{0.785}\tabularnewline

Proposed Fast-VVLAD & $128 \times 13$ & 6.55 & 6.66k & 0.490 & 0.745 & 0.728\tabularnewline

Multi-VLAD \cite{arandjelovic2013all}& $128 \times 14$ & 14 & 7.17k & 0.493 & \textbf{0.780} & 0.732\tabularnewline

\hline
\end{tabular}
}
\label{tab:mAPCars}
\end{table}

\begin{table}
\caption{{Complexity and mAP\ results  for the Holidays dataset \cite{jegou2008hamming}}. }
\centering
\resizebox{\columnwidth}{!}{
\begin{tabular}{cc>{\centering}p{1.5cm}>{\centering}p{1cm}>{\centering}p{1cm}>{\centering}p{1cm}}
\hline\noalign{\smallskip}
 & $D_\text{tot}$  &  Matching Complexity & Descriptor Storage (bytes) & Query ROI & Whole Image\tabularnewline
\noalign{\smallskip}
\hline
\noalign{\smallskip}
CNN (Layer 13)   & 128 & 1 & 512 &0.339  &0.757\tabularnewline

& 512\ & 4 & 2.05k& 0.369 &\textbf{0.767}\tabularnewline

Proposed Fast-VDCNN  & $128 \times 13$ &  6.11 & 6.66k  &0.674  &0.761 \tabularnewline

Multi-CNN \cite{arandjelovic2013all,chatfield2014return}  & $128 \times 14$ & 14 & 7.17k  &\textbf{0.678}  &0.737\tabularnewline
 
\hline
\end{tabular}
}

\label{tab:mAPHol}
\end{table}

\noindent\textbf{Caltech + Stanford Cars:}\ Table \ref{tab:mAPCarsQ} summarises the performance of the various descriptors with WNPQ on the Caltech + Stanford Cars dataset.  On whole images, coupled with the aggregated similarity score, Fast-VVLAD\ offers superior performance to the 128-D VLAD, with an mAP gain of 6\%.  The 1664-D VLAD, which is now of comparable complexity to the Fast-VVLAD, is outperformed on  the small license plate queries, with mAP gain of 9\%, but remains superior for whole-image queries. However, it is worth mentioning that the gain from Fast-VVLAD on small queries outweighs any loss on larger queries, thus making it favorable. Finally, the quantized Multi-VLAD offers marginally superior mAP to Fast-VVLAD, albeit at the cost of twice the matching complexity and higher descriptor storage size\footnote{It is worth noting that, given we use a single PQ codebook for quantizing all cell components of Fast-VVLAD and Fast-VDCNN, all quantized  based systems\ have a $Z' \times Z '\times M$ bit cost for storing the look-up tables.  This means that, for example, although the 1664-D VLAD offers a lower storage size to Fast-VVLAD, there is an additional 1.7MB cost to store the look-up table, versus 262kB for Fast-VVLAD. However, as mentioned previously, the significance of this additional storage cost diminishes when increasing the test dataset size.}.  \\

\noindent\textbf{Holidays: }For the Holidays dataset,  the quantized Fast-VDCNN maintains its mAP gain on query ROI over the quantized 128-D and 512-D  CNN feature descriptors, while the descriptor storage  has been reduced by a factor of 16 compared to its unquantized counterpart. In addition,  the Fast-VDCNN still performs better than quantized Multi-CNN\  on whole image queries, with an mAP gain of 4\%.\ \  \

\begin{table}
\caption{Complexity and mAP\ results for the Caltech + Stanford Cars image dataset with WNPQ\cite{fergus2003object, krause20133d}.}
\centering
\resizebox{\columnwidth}{!}{
\begin{tabular}{cc>{\centering}p{1.5cm}>{\centering}p{1cm}>{\centering}p{1cm}>{\centering}p{1cm}>{\centering}p{1cm}}
\hline\noalign{\smallskip}
 & $D_\text{tot}$  &  Matching Complexity & Descriptor Storage (bytes) & License\\ Plates & Trunk & Whole Image\tabularnewline
\noalign{\smallskip}
\hline
\noalign{\smallskip} 
WNPQ VLAD   & 128 & 1 & 32 & 0.112 &0.606 &0.626\tabularnewline

& 768  & 3 & 96 & 0.257 & 0.663 & 0.677\tabularnewline

& 1664 & 6.5 & 208 & 0.404 &0.696 &\textbf{0.702}\tabularnewline

Proposed WNPQ Fast-VVLAD & 128 $\times$ 13 & 6.43 & 416 & 0.440 & 0.713 & 0.661\tabularnewline

WNPQ Multi-VLAD & 128 $\times$ 14 & 14 & 448 & \textbf{0.449} & \textbf{0.769} & 0.652\tabularnewline

\hline
\end{tabular}
}
\label{tab:mAPCarsQ}
\end{table}

\begin{table}
\caption{Complexity and mAP\ results  for the Holidays dataset with WNPQ \cite{jegou2008hamming}.}
\centering
\resizebox{\columnwidth}{!}{
\begin{tabular}{cc>{\centering}p{1.5cm}>{\centering}p{1cm}>{\centering}p{1cm}>{\centering}p{1cm}}
\hline\noalign{\smallskip}
 & $D_\text{tot}$  &  Matching Complexity & Descriptor Storage (bytes) & Query ROI & Whole Image\tabularnewline
\noalign{\smallskip}
\hline
\noalign{\smallskip} 
WNPQ CNN (Layer 13)   & 128& 1 & 32  & 0.248  &0.674\tabularnewline
 & 512$$  & 4 & 128  & 0.280 & \textbf{0.706}\tabularnewline

Proposed WNPQ Fast-VDCNN &128 $\times$ 13 & 6.13 & 416 & 0.550 & 0.684\tabularnewline

WNPQ Multi-CNN &128 $\times$ 14 & 14& 448 & \textbf{0.603} & 0.656\tabularnewline
 
\hline
\end{tabular}
}

\end{table}

\subsection{Further Improvements on Whole-image Search} \label{compHol}

The experimental results of the previous section show that Fast-VVLAD and Fast-VDCNN clearly outperform their counterparts for ROI image search, while being competitive for whole-image search. The performance on whole image queries is primarily controlled by the dimension of the level-0 (whole image)  component. {For experiments in the previous section, we set the dimension uniformly across all components of the Voronoi-based descriptor, i.e., 128-D descriptor per cell. As a result, mAP for the Voronoi-based descriptors\ on whole images is comparable to that of the 128-D  reference descriptors.  One option to tailor performance towards whole image queries or smaller ROI queries is by tapering the dimension across levels; we leave this as a topic for future study.}

{Another approach to boost performance for whole image queries is by accounting for multiple scales in \textit{both} the query and dataset images. In other words, rather than applying Voronoi partitioning \textit{only} on the dataset images, we can also apply Voronoi partitioning \textit{on the query image} over multiple levels and submit each of the $V_\text{tot}$ query partitions as a subquery. Notably, using the Fast-VDCNN for the dataset image encodings, each subquery  is matched only against representative cells in the dataset images (i.e., between 4 to 7 cells), which are determined by the adaptive search proposed in Section \ref{Fast-VVLAD}.  The inner product between the original query image and a dataset image is taken as the average inner product over all $V_\text{tot}$ subqueries.} While this incurs linear increase in the search complexity (by   $V_\text{tot}$), this scales better than the quadratic search complexity achieved by Carlsson \textit{et al.} \cite{razavian2014cnn, azizpour2014factors}, where exhaustive search amongst all subqueries is carried out.   

{Table \ref{tab:mAPHolWhole} compares the retrieval performance of the proposed Fast-VDCNN descriptor against the current state-of-the-art on the Holidays dataset that use networks pre-trained on ImageNet. {The Fast-VDCNN descriptor is generated under the configuration of Section \ref{expDet}, albeit now also partitioning the queries with $V_1 = V_2 = 2$ and resizing image partitions to $448 \times 448$}. Beyond benchmarking against the grid-based spatial search method of Carlsson \textit{et al.} \cite{razavian2014cnn}, we also compare our results with the recently-proposed CNN+VLAD \cite{gong2014multi}, CKN-mix \cite{paulin2016convolutional}, the hybrid FV-NN approach of Peronnin \textit{et al.} \cite{perronnin2015fisher}, as well as lower-dimensional but more computationally-intensive proposals \footnote{{In particular: the SPoC descriptor \cite{babenko2015aggregating} offers the best performance to dimensionality, but utilizes a {deeper and a more computationally heavy CNN  (144M\ parameters vs 76M parameters for our architecture)} and a larger image input size, the R-MAC based\ descriptor  uses Siamese learning with supervised whitening, and NetVLAD\ requires additional processing (soft assignment and normalizations within the NetVLAD layer) to encode VLAD from the network activations. On the contrary, under the chosen configuration, the proposed Fast-VDCNN approach allocates only 128 dimensions per cell and accesses between 4 to 7 cells for each image subquery. }}  that perform competitively \cite{babenko2015aggregating,radenovic2016cnn,arandjelovic2015netvlad}.  Evidently, the additional scale and location invariance provided by the  Voronoi partitioning leads to the proposed Fast-VDCNN achieving competitive performance to other CNN  derived frameworks and hybrid variants, {without manually rotating the images,} and despite the fact that our feature descriptor is built directly from a pre-trained network  and incurs modest computational and storage requirements.   }

\begin{table}
\caption{Comparison of whole-image retrieval performance (mAP)\ with state-of-the-art  for the Holidays dataset \cite{jegou2008hamming}. The proposed approach allocates 128 dimensions per partition cell.} 
\centering{
\begin{tabular}{cc>{\centering}p{1.5cm}>{\centering}p{1cm}>{\centering}p{1cm}>{\centering}p{1cm}}
\hline\noalign{\smallskip}
 & $D_\text{tot}$   & Whole Image\tabularnewline
\noalign{\smallskip}
\hline
\noalign{\smallskip}
Proposed Fast-VDCNN  & 1.66K (128)  &  {0.821}
\tabularnewline
FV-NN (Peronnin \textit{et al.}) \cite{perronnin2015fisher}    & 4K  &\textbf{0.835}
\tabularnewline
CNN + VLAD \cite{gong2014multi}  & 2K  & 0.802  \tabularnewline

CNN (Carlsson \textit{et al.}) \cite{razavian2014cnn} & 4K-15K  & 0.769  \tabularnewline

CKN-mix \cite{paulin2016convolutional} & 4K & 0.829 \tabularnewline

SPoC (w/o center prior) \cite{babenko2015aggregating}  & 256 & 0.802 \tabularnewline

R-MAC \cite{radenovic2016cnn} & 512 & 0.825\tabularnewline

NetVLAD \cite{arandjelovic2015netvlad} & 256 & 0.799 \tabularnewline 
\hline
\label{tab:mAPHolWhole}
\end{tabular}
}
\end{table}

 





\section{Conclusion}
\label{sec:Conclusion}
We proposed a novel descriptor design, termed Voronoi-based encoding,\  for region-of-interest image retrieval. We have shown how VE could fit into a practical ROI-based retrieval system via the proposed fast search, memory-efficient design, product-quantization based\ lossy compression techniques, and robust similarity scoring mechanisms. We test retrieval performance on two datasets, using VLAD\ and a deep CNN\ as our descriptor basis. Our results show that our approach is descriptor agnostic;  the proposed  Fast-VVLAD\ and Fast-VDCNN  maintain competitive retrieval performance over diverse ROI queries on two datasets and significantly improve on the retrieval performance (or implementation efficiency)\ of their respective descriptor variants\  with a grid spatial search, when dealing with smaller ROI queries.
Moreover, improved geometric invariance  results in competitive retrieval performance to the current state-of-the-art on whole image queries.


%

\appendices
\section{Level Projection for VE Storage Compaction} \label{App A}
In order to decrease the storage requirements for unquantized Voronoi-based encoded (VE) representations, the  descriptor over two constituent cells \(x\) and \(y\) (i.e., spatially-neighboring cells belonging to the same cell of the upper level), can be approximated as: 
\begin{equation}
 \widetilde{\boldsymbol{\phi}}_{x \cup y} = \widetilde{\boldsymbol{\phi}}_{x} + \widetilde{\boldsymbol{\phi}}_{y}.\label{phi_union}\end{equation}
This holds because both PCA and whitening are linear mappings, therefore, if we do not consider the vector truncation and subsequent $L_2$ normalization  of the individual cell\  vectors, the additivity property holds in the projected domain as well.
 Given that directionality is preserved under normalization,  \eqref{phi_union}  provides an approximation to the normalized encoding\ computed directly over the two cells.   Therefore, we can trade-off computation for memory by solely storing the last-level PCA-projected descriptors (level $L-1$) and computing all other cell encodings\ for all lower levels at runtime via repetitive application of \eqref{phi_union} amongst constituent cells and renormalizing before carrying out the similarity measurement of \eqref{similarity}. This is an appealing proposition for practical systems because vectorized addition and scaling for normalization is extremely inexpensive in modern SIMD-based architectures.
As such,  this approach requires storing only   $\prod^{L-1}_{l=1} V_l$ cell descriptors,   instead of  \(V_\text{tot}\) cell\ vectors. Naturally, there is a dependency on the projection error, which will evidently be greater with less dimensions retained post-PCA.    

We can integrate product quantization with a modified level projection for quantized VE storage compaction.  As before, we only store the last-level (quantized) descriptors offline.  However, as the inner product satisfies the distributive law, we should now directly approximate the inner product between a query encoding and a level $l-1$ cell descriptor as an $L_1$ normalized summation:

\begin{equation}
S_{\text{ROI},l-1}= \frac{1}{V_l}\sum^{V_l}_{i=1}S_{\text{ROI},l_i \in l-1}
\end{equation}    
where each inner product is read from a look up table. 

\section{Proof of Proposition \ref{prop1}}\label{App_b}
\begin{IEEEproof}   
In order to optimize the   bit allocation to the various descriptions (subspaces), we optimize the rate distortion expression:
\begin{equation}
R(E) = \min_{\sum{E_m=E}}\sum^{M}_{m=1}\ln\frac{\left\vert \mathbf{\Sigma}_m\right\vert}{E_m}
\end{equation}
where $E_m$ is the product of the subcomponent distortions,  $\left\vert \mathbf{\Sigma}_m \right\vert$ is the determinant of the covariance matrix $\mathbf{\Sigma}_m $ and $E$ is the overall distortion value.  The minimum rate for given $E$ is derived when all distortions $E_m$ are equal, i.e., $E_m=E/M$.

Using results derived from rate distortion theory \cite{gersho2012vector},  $E_m$ for the $m$-th subspace can be approximated by: 

\begin{equation}
E_m\approx\ \left\vert \mathbf{\Sigma}_m \right\vert \prod^{D'}_{i=1}h_{im}2^{-2b_{im}}=\left\vert \mathbf{\Sigma}_m \right\vert h'_{m}2^{-2B'_m}
\end{equation}
where $h_{im}=\frac{1}{12}\left\{ \int^\infty_{-\infty}f_{im}\left({x}  \right)\ dx\right\}^3$ is a variable determined by the univariate Gaussian of the normalized components, $f_{im}\left( x \right)$, and $b_{im}$ is the average number of bits encoded per dimension.  Due to the independence property, the product of $h_{im}$ in the $m$-th subspace yields the variable  $h'_m$, which is now determined by the   multivariate Gaussian distribution for the normalized subspace random vectors.   This distribution is independent of subspace, and, as such,  $h'_m$ is constant for all $m$.  Similarly, if the size of the bit encoding and block dimension $D'$ is fixed per subspace, then   $B'_{m}$ is  a constant for all $m$.   For $E_m$ to be equal for all $m$, \(\left\vert \mathbf{\Sigma}_m \right\vert\) must be constant, independent of subspace.
\end{IEEEproof} 


%
%

\ifCLASSOPTIONcaptionsoff
  \newpage
\fi



\bibliographystyle{IEEEtran}
\nocite{*}
\bibliography{ITMM_v26_arxiv}

%
%

\begin{IEEEbiography}[{\includegraphics[width=1.1in,height=1.25in, clip, keepaspectratio]{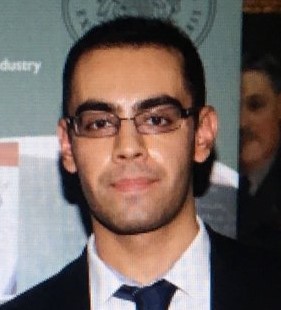}}]{Aaron Chadha} obtained a BA and MEng in Information and Computer Engineering from the University of Cambridge, graduating with a Distinction.  He joined the Department of Electronic and Electrical Engineering at University College London in 2014, where he is currently working towards a PhD in the area of image/video classification and retrieval under the supervision of Dr Yiannis Andreopoulos.  He was awarded with an Industrial Fellowship from the Royal Commission of the Exhibition of 1851 in 2016.  \end{IEEEbiography}

\begin{IEEEbiography}[{\includegraphics[width=1.1in,height=1.25in, clip, keepaspectratio]{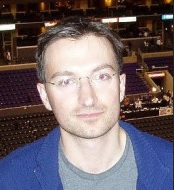}}]{Yiannis Andreopoulos}(M'00--SM'14) obtained the Electrical Engineering Diploma and an MSc in Signal and Image Processing Systems from the University of Patras, Greece, and the PhD\ in Applied Sciences from the Vrije Universiteit Brussel, Belgium. He is Reader in Data and Signal Processing Systems in the Department of Electronic and Electrical Engineering of University College London (U.K.). His research interests are in wireless sensor networks, error-tolerant computing and multimedia systems. He received the 2007 Most-Cited Paper Award from the Elsevier EURASIP Signal Processing: Image Communication journal and a best paper award from the 2009 IEEE Workshop on Signal Processing Systems. He was Special Sessions Co-Chair of the 10th International Workshop on Image Analysis for Multimedia Interactive Services (WIAMIS 2009) and Programme Co-Chair of the 18th International Conference on Multimedia Modeling (MMM 2012) and the 9th International Conference on Body Area Networks (BODYNETS 2014). He has been an Associate Editor of the {IEEE Transactions on Multimedia}, the IEEE Signal Processing Letters and Image and Vision Computing (Elsevier). \end{IEEEbiography}








\end{document}